# MODEL INTERPRETATION AND EXPLAINABILITY
## Towards Creating Transparency in Prediction Models

| Donald Kridel | Jacob Dineen | Daniel Dolk | David Castillo |
| U Missouri, St Louis | Univ. of Virginia | Naval Postgrad School | Capital One, Inc. |
| dkridel@gmail.com | jdineen81294@gmail.com | drdolk@nps.edu | dcastilloaz@gmail.com |

**Abstract**

*Explainable AI (XAI) has a counterpart in analytical modeling which we refer to as model explainability. We tackle the issue of model explainability in the context of prediction models. We analyze a dataset of loans from a credit card company using the following three steps: execute and compare four different prediction methods, apply the best known explainability techniques in the current literature to the model training sets to identify feature importance (FI) (static case), and finally to cross-check whether the FI set holds up under "what if" prediction scenarios for continuous and categorical variables (dynamic case). We found inconsistency in FI identification between the static and dynamic cases. We summarize the "state of the art" in model explainability and suggest further research to advance the field.*

## 1. Introduction and Background

Given the recent success of machine learning algorithms (MLAs) and the attendant angst surrounding the potential negative impact of AI on our society [23], *explainable AI* has now become an area of increased scrutiny and research. Since MLAs, especially neural networks, tend to be "black boxes" and highly nonlinear in nature, it is often not clear, even to experienced practitioners, how particular decision outcomes are reached. This, in turn, leads to a vague apprehension that MLAs may soon outstrip human ability to understand and manage their results. Without addressing this existential concern explicitly, we tackle here a more focused and pragmatic dimension of the problem, namely how to interpret and explain prediction models.

Explainability and interpretation are problems which have plagued analytical models as well. For example optimization and advanced econometric models have typically met with significant resistance from management decision makers for whom they have been designed. Translating mathematical expertise into decision-making expertise still remains a significant obstacle in gaining management acceptance of model artifacts. It is not unreasonable to expect that advances in model explainability and interpretation can help bridge this gap.

Model explainability and interpretability are now being perceived as desirable, if not required, features of data science and predictive analytics overall. Our objective here is to examine what these features may look like when applied to previous research we have conducted in the area of econometric prediction and predictive analytics [10]. We consider the domain of Lending Club loan applications. For our dataset, we perform three different analyses:

1. Model Execution and Comparison. Run and compare four different prediction models on the training set as shown in Table 1 (logistic regression, random forest, boosted gradient, multi-layer perceptron (MLP neural network));
2. Explainability Model Execution and Comparison (training dataset only). For each model, apply existing model explainability techniques (Local Interpretable Model Explanation (LIME), SHAP (SHapley Additive exPlanations), GAM, and SKLearn to the static training dataset in order to assess the comparability of these approaches with respect major feature identification.
3. What-if or Perturbation Analysis. In the 3rd and final step, we examine how well the explainability models hold up under dynamic prediction situations wherein we perturb the major features identified in Step 2 and compare the explainability models to the static (training set) case.

Most predictive model explainability approaches focus on the static part of the process whereas our contribution is to identify a more general approach to prediction model explainability for decision makers that holds up under both static and dynamic scenarios.

## 2. Review of Selected Explainability Approaches to Prediction Models

Several techniques have been developed to address the problem of explainable predictions. Broadly speaking, these techniques employ various forms of sensitivity analysis to identify a streamlined *feature importance set* (also called *feature attribution*) having the greatest impact upon a prediction. These procedures vary depending upon how they measure the impact of a feature upon a specific local (point) and/or global prediction.

To differentiate local vs global explainability, consider a prediction model which targets customers who may be inclined to respond to a specific marketing campaign. Local explainability is customer-specific, that is, it purports to explain what features, or attributes, influenced an individual customer to respond (or not) to the campaign ad(s). Global explainability on the other hand would try to identify a set of salient features which influenced all customers who responded. The latter would clearly be useful in designing future marketing campaigns.

**Table 1. Prediction models generated for explainability application**

| Analytical Method | Refs | Description |
| --- | --- | --- |
| Logistic regression (Logit) | [24] | Discrete choice regression |
| Random Forest (RF) | [3,5,8] | Random Forest is a supervised learning algorithm which builds and merges multiple decision trees to obtain an accurate and stable prediction. |
| Gradient Boosting (GBC) | [6] | Machine learning technique for regression and classification problems, which generates a prediction model as an optimization of a loss function across an ensemble of weak prediction models, typically decision trees. |
| MLP Neural Net (N/N) | [16] | Implementation of Deep Neural Networks |

Recently, explainability techniques have been proliferating rapidly in response to the perceived need to render deep learning algorithms more transparent [15]. However,, there has been research in the past which explores the accuracy of model transparency. For example, [2] uses Interactions-based Method for Explanation (IME) [21] and EXPLAIN [19] to determine feature importance. IME computes feature importance by dropping a single feature and measuring contribution, whereas EXPLAIN processes permutations of subsets, iteratively dropping n features and measuring the resultant contributions. A weighted distance equation is then generated in order to compare the explanations of support vector machines (SVM), artificial neural nets (ANN), and k-nearest neighbors (KNN) to the learned structure of a decision tree to remove the subjectivity of the explanations globally.

[7] approaches explainability and fairness in AI, from a philosophical perspective which intersects with our core message of the need for explainable predictions in industry. They discuss how nonlinear function approximators (Boosting / Bagging / Neural Nets) suffer from some issues of explainability due to the summation of multiple classifiers, the use of voting classifiers, hidden layers and activation function. They don't discuss current "state of the art" in explainability, but rather ponder the overall pipeline of data collection, model construction, and model use.

Our approach is more specific and closely aligned with recent explainability techniques shown in Table 2, which we chose according to the criteria:
1. Techniques must be "model agnostic" and thus readily adaptable to classifier- and regression-based prediction applications.
2. Techniques must have available Python code accessible from *GitHub* or equivalent sources. This relieves us from having to develop N/N-based prediction models as well as writing code to implement explainable model algorithms.

**Table 2. Model explainability techniques to be applied to models in Table 1.**

| Explain-ability Technique | Refs | Brief Description |
| --- | --- | --- |
| SKLearn Feature Importance | [13] | SKLearn library |
| LIME (Local Interpretable Model-agnostic Explanations) | [17,18] | Generates linear approximations to a model by random sampling in a local neighborhood and fitting a simpler linear model to the newly constructed synthetic data set. |
| SHAP (SHapley Additive exPlanations) | [11] | An *additive feature attribution* method that generates a linear explanation model assigning an importance value to each feature reflecting its effect on the model. |
| GAM (Global Attribute Model) | [9, 14] | GAM has a global vs. local focus, grouping similar local feature importance to form human-interpretable global attributions that best explain a particular subset of the data. |

## 3. Process (Multiple Models Applied to Dataset)

For our dataset, we sample 20,000 observations, enforcing class balance, from Lending Club's publicly sourced dataset (pertaining to active and past loans) [1]- Active and past loans that have been fully paid or have no existing derogatory marks are classified as 'good loans'. Conversely, 'bad loans' are instances where an individual has either defaulted or is currently delinquent. What we want to predict is whether an individual loan is "bad" (BadLoan vs. GoodLoan) because of factors such as payment defaults, late payments, high balances, etc.

We start by running 4 standard models[2]:
1. Logistic- this is the reference model due to its "easy" explainability [24]
2. Random Forest [3,5,8]
3. Boosted Gradient [4,6]
4. Neural Network [16]

Logistic regression is widely-used in industry (and has been for several decades); random forest, and gradient-boosted classifier are popular tree-based ML techniques. For neural networks, we consider two estimation options: a simple SKLearn-estimated neural net and a richer neural net utilizing KERAS/TensorFlow. The SKLearn neural network is a binary classification network with a single hidden layer consisting of 150 neurons. This was, for the most part, an 'out of the box' classifier. We also train a multilayer perceptron (MLP) binary classification network with four hidden layers of arbitrary depth, utilizing batch normalization and probabilistic dropout for regularization. The network uses the rectified linear unit (relu) activation and optimizes based on cross-entropy loss with a variant of stochastic gradient descent (Adam). Since we only need one N/N for comparison purposes and the MLP model is more robust (Table 3), we will not consider the SKLearn prediction model further in our analyses. We will however still be using SKLearn as an explainability technique separate from its application as a prediction technique.

Our 1st step is to compare the remaining 4 models with respect to how well they predict the classifier. Table 3 shows the comparative accuracy of the predictions with respect to the class attribute (good loan [+] or bad loan [-]).

**Table 3. Prediction accuracy for each model**

| Model Type | Prediction Accuracy |
|---|---|
| Logistic Regression | 88.4% |
| Random Forest | 90.0% |
| Gradient Boosting | 94.1% |
| MLP Neural Net | 87.7% |
| Simple (SKlearn) Neural Net | 83.1% |

## 4. Comparison of Explainability Techniques

**A. SKLearn** provides a standard library for identifying feature importance most often used on tree-based classifiers. These feature importance measures can be based on gini importance (mean decrease impurity) or mean decrease accuracy. Since SKLearn is widely used, we begin our 'importance' measures here.

Figure 1 details the SKLearn feature importance for the Random Forest (RF) model. This is a typical representation format for easy visualization of the relative impact of features, or attributes, on a prediction model.

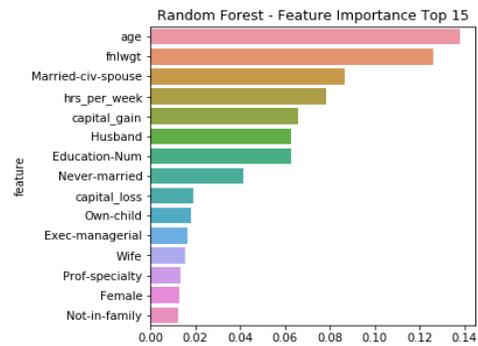

**Figure 1. Feature importance for random forest prediction as determined by SKLearn**

We did not make a concerted effort to optimize any of the models as we might, if we were to actually deploy one of these models, since our objective is to examine the explainability metrics across the models rather than the prediction accuracy for any specific model as in a usual deployment scenario.

---

[1] Reference link to dataset:
https://www.kaggle.com/wendykan/lending-club-loan-data#LCDataDictionary.xlsx
[2] The software suite used for the analytics described in this paper consists of the Anaconda environment, Jupyter notebook, keras, tensorflow, various algorithms available via GitHub (e.g., SHAP and GAM), and the Python programming language

**B. LIME** (Local Interpretable Model-agnostic Explanations) [17, 18]: LIME (and related techniques SP-LIME and aLIME) generates linear approximations to a model by random sampling in a local neighborhood and fitting a simpler linear model to the newly constructed synthetic data set. The now explainable linear model's weights can be used to interpret a particular (i.e., local) model prediction. This method can be applied to neural networks or any uninterpretable nonlinear model and is thus described as *model agnostic*. LIME is particularly useful for local interpretability but can be applied globally by summing all the individual point explanations. Although LIME was the first model explainability technique to appear in the literature, SHAP and GAM claim to be more general techniques that subsume LIME. As a result (and because of space limitations), we will not consider global LIME here.

**C. SHAP** (SHapley Additive exPlanations) [11, 12] The Shapley Value (SV) has its genesis in game theory where SV represents each player's input over all possible combinations of players. This approach yields a model called the Shapley Value regression [11]. SHAP is an *additive feature attribution* method that generates a linear explanation model whose regression values are feature importance values for linear models in the presence of multicollinearity. This method assigns an importance value to each feature that represents the effect on the model prediction of including that feature. To compute this effect, a model is trained with that feature present, and another model is trained with the feature withheld and the impact difference on the prediction is then measured.

Figure 2 shows a typical Shapley display graph for the Logistic Regression model. Each dot on the horizontal axis represents a row of the dataset with blue dots representing low values and red dots high values. The feature attribution rankings (top to bottom in Figure 2) are based on Rank = $\sum(|shap\_score|)$ so the first feature has the highest sum of absolute shap scores. Shapley graphs provide a clear ranking of feature importance, but can be displayed more intuitively as Feature Importance graphs by summing the absolute SHAP-scores (Figure 3).

While the order is slightly different, nine of the ten most important features (calculated directly from the coefficients from the logit model) are in the SHAP Top 10 (Figure 2). The "missing feature" from the SHAP top-10 is initial payment (it is 10[th] in importance with direct calculation and is 18[th] in SHAP calculation); Loan-grade B is ranked 10[th] in SHAP while it is 13[th] in direct calculation). The SHAP importance rankings seem quite consistent with the 'true values' for the logit classifier

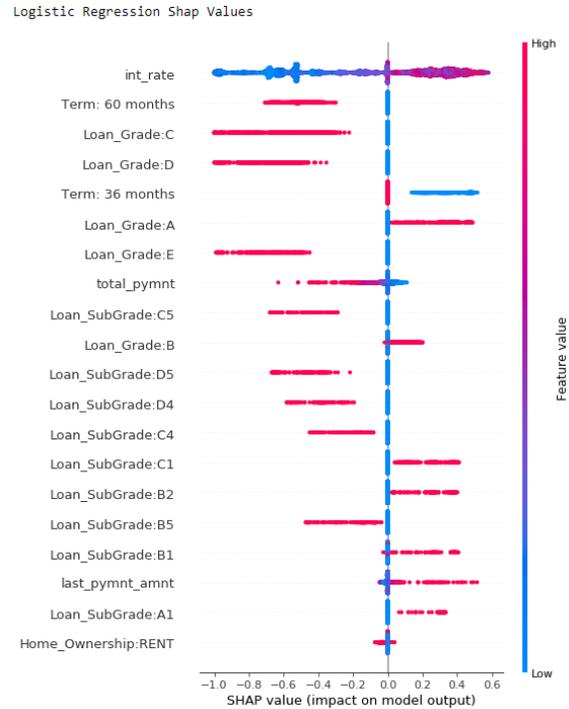

**Figure 2. Shapley graph for logistic regression model**

.The Shapley Feature Importance graphs in Figure 3 show considerable overlap suggesting that the same features are generally important in each of the models. In particular, the attributes *total_pymnt* appears in the top-10 for all four models, and the *int_rate* appears very important (except in the MLP). The logit model, perhaps due to its linear index (between the choices), has more categorical features (loan-types) in its most important features.

**D. GAM** (Global Attribute Model) [9, 14]. GAM explains the landscape of neural network predictions across subpopulations. GAM augments global explanations with the proportion of samples that each attribution best explains and specifies which samples are described by each attribution. The advantages of GAM's global explanations 1) yield the known feature importance of simulated data, 2) match feature weights of interpretable statistical models on real data, and 3) are intuitive to practitioners through user studies.

We run GAM on a subsample of 1000 attribution values (on the MLP neural network) for each class for both balanced and unbalanced subpopulations. We then forced our subpopulations to explicitly map to our class labels to provide explanations for the target variable: one for the GoodLoan group and the other for BadLoan group, as shown in Figure 4. As the figures (ranked by feature importance in descending order) show, last_pyment_amnt has the highest Feature Importance for the GoodLoan subgroup and the BadLoan subgroup indicating that it is a critical attribute in predicting loan defaults.

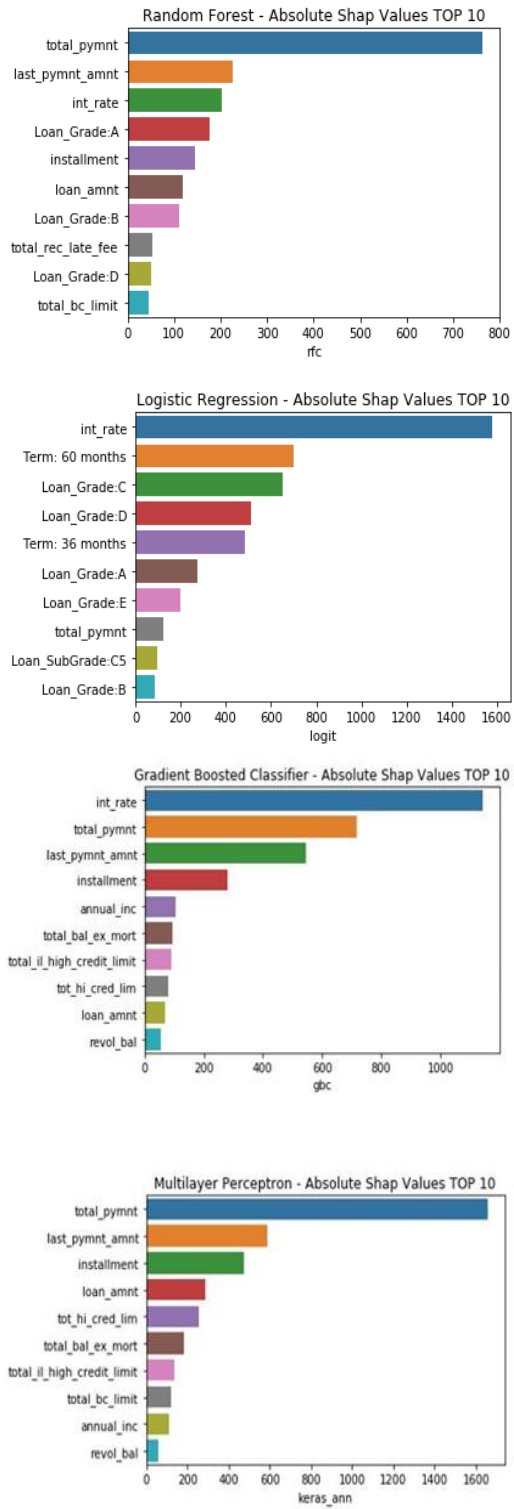

**Figure 3. Shapley feature importance graphs for each model**

It should be mentioned that there are additional explainability techniques not considered here, for example DeepLIFT [20] and Integrated Gradients [22] are both examples of gradient-based methods [1] and are primarily used in image recognition applications. While a complete analysis of all explainability methods is beyond the scope of this paper, we mention DeepLIFT and Integrated Gradients because they are popular techniques for Deep Learning (N/N) models. The explainability methods we have chosen for our analysis have specific relevance to prediction models, but also can, in principle, be applied across broad classes of models including Deep Learning (N/N) models. As we indicate in our future research discussion, we intend to expand our analyses to include a wider range of these explainability techniques.

The previous techniques help shed some light on the prediction model black box by giving us a sense of feature importance for the training set utilized to develop the models. Feature Importance graphs highlight the major influencers and allow us a more or less intuitive grasp of where to focus our attention. For example, we can see from Figures 3 and 4 that the features *int_rate*, *last_pymnt_amnt*, and *total_pymnt* play prominent roles across (most) of the estimated models. This can, at a minimum, serve as a basis for more detailed drill down analysis. We should mention that we are not necessarily looking for consensus across models but rather we want to know whether feature importance metrics allow us to gauge the impact on predictions. We now turn our attention to the dynamic case where we use the models to make predictions and examine Feature Importance in that context.

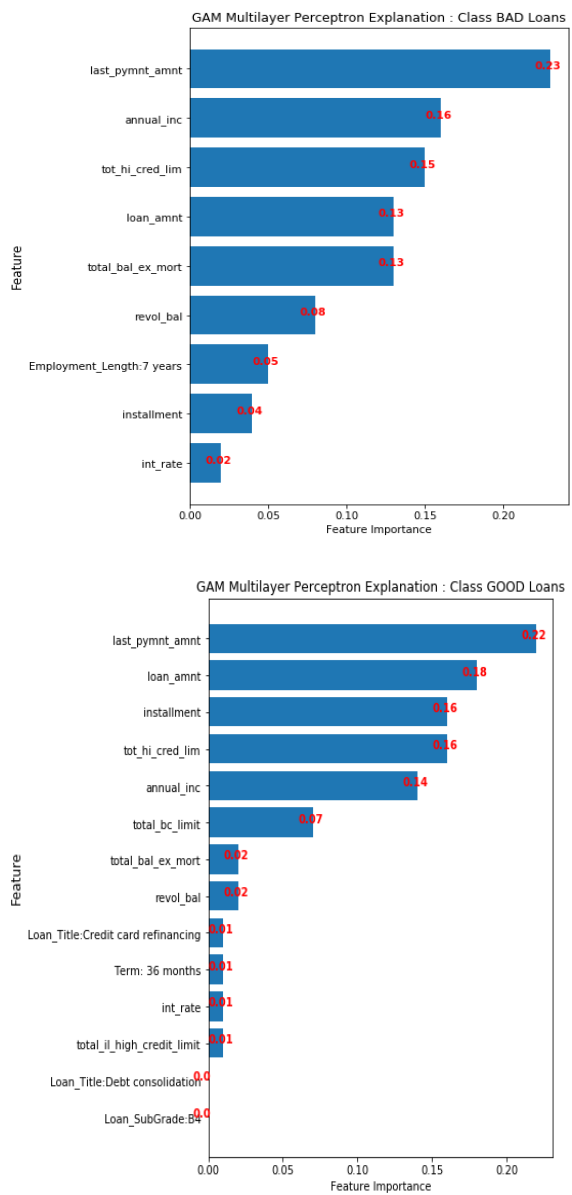

**Figure 4. GAM explainability plots for loan application data set.**

## 5. Prediction

We are interested in the case where the model will be utilized for prediction. For example, a common use-case would be in determining whether or not an applicant for a credit-card (or a loan) should be "accepted". If the answer is to accept the applicant (grant the 'loan' request), then an explanation to the may be useful for regulatory oversight and accountability. However, in the case of rejecting an applicant (at least in the US), the bank is required to provide the applicant with reasons for refusing the application.

These 'explanations' would be provided not only to the would-be customer but also to bank personnel who must interact with the applicant and to credit rating organizations. Further, in many cases, suggestions for behavioral changes must be made to the applicant so that they will have a higher chance of acceptance in any subsequent applications. Typically, "reason codes" are developed from the scoring models (often logit models) and these reason codes provide the basis for these explanations (and remediation suggestions). As a result, both global (for model governance approval) and local (for individual predictions) explanations are not only useful but often required.

In the current data-set we have utilized, we predict which loans will be "Bad" and which "Good". Since there are no "new applications" available, the approach we have taken, is to perform some simple "what if" perturbation analysis on a hold-out sample. We can then compare sensitivity of these predictions to the 'feature importance' results in the previous section. This will allow us to see whether our feature importance hypotheses hold up equally well in a prediction scenario compared to the standard training data case. The process for perturbing continuous variables and categorical variables will vary slightly.

For continuous features, we choose to focus on features that come across the importance horizon: the interest rate (*int*) which tended to have high importance, income (*ann_income*) which had low importance, and payment (*total_pymnt*) which was mixed.

Our approach is to run multiple "what if" scenarios (between 0.5 and 1.5) vis-à-vis the base case, tweaking one feature while holding the others constant. Ideally, we would like to see the sensitivity in the prediction scenarios mirror the feature importance suggested for the training set by the explainability techniques. Figures 5A,B and Table 4 show the results of our perturbations for the continuous variables. Not surprisingly, logit prediction sensitivity follows expectations since model parameters are explicit. The interest rate is the most sensitive; in fact, it is likely too sensitive and if the model were to be deployed more development would be required. Further, the next two most sensitive are total and last payments.

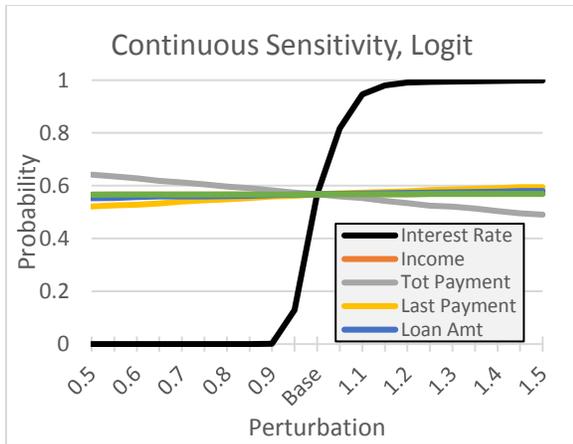

**Figure 5A. Logit prediction sensitivities for perturbed continuous variables. (logit and random forest)**

Sensitivities for the RF are 'generally' consistent with SHAP-importance values. The predictions are most sensitive to the interest rate (the 3$^{rd}$ highest SHAP score) with predictions being second-most sensitive to total payment (highest shap score). Model predictions are quite insensitive to *last_payment_amount* (second highest SHAP score). Note also the "peculiar shape" of predictions for the interest rate: decrease in the interest rate lead to decreased probabilities, but increases in the interest rate also lead to decrease in the probabilities. Such a "sign change" would be difficult to explain and would likely prevent model use in highly-regulated industries.

Like the RF sensitivities, the GBC findings are again broadly consistent with expectations based on the SHAP importance values. GBC model predictions are most sensitive to the *interest rate* (the highest SHAP score) with predictions being second-most sensitive to *last payment* (third highest SHAP-score). Model predictions are quite insensitive to *Tot Payment* (second highest SHAP-score).

Note also the "peculiar shape" of predictions for the both interest rate and total payment. There are several 'sign reversals' in the interest rate projections (though smaller than was observed in RF). For *Tot Payment*, response is very flat for reductions, but large (and incorrectly signed) for increases. Once again these "sign issues" would almost surely become real-world deployment issues.

For the MLP NN, we see consistent results—in terms of agreement between sensitivity and feature importance. Model predictions are most sensitive to total payment; second most sensitive to last payment and third most sensitive to installment. Total payment has the highest SHAP-score but does not appear in the top GAM scores (for GoodLoan subpopulation). Last payment has the second highest SHAP-score and was ranked as the most important feature by GAM. Installment was third in both SHAP and GAM. We do see a small "sign reversal" of the marginal impact for total payments (around .5); this, however, is much smaller than what was observed for the RF and GBC models.

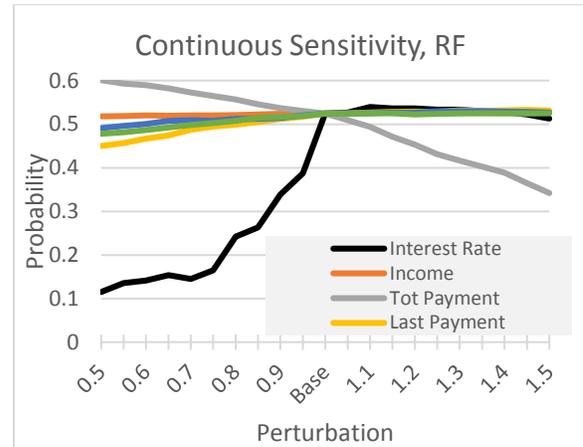

**Figure 5B. Random forest (RF) prediction sensitivities for perturbed continuous variables.**

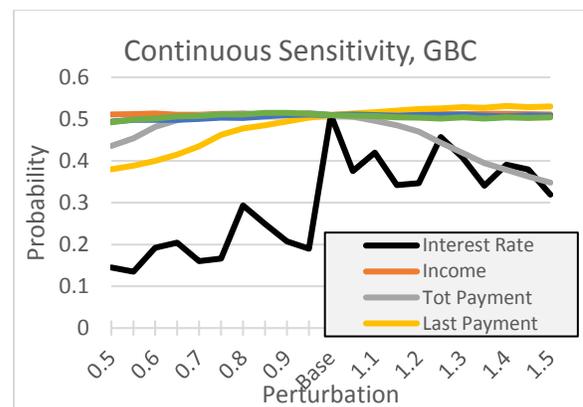

**Figure 6A. Gradient-boosted classifier (GBC) prediction sensitivities for perturbed continuous variables.**

For the categorical features, we consider loan grade and loan title. Note that each category value results in a different independent variable (hot encoded). In this case, we randomly select 0's of a specific category, change some of these (in increasing proportions) to 1's and measure impact on predicted probabilities for these changes.[3] Table 5 and Figure 7 shows the results of these perturbations

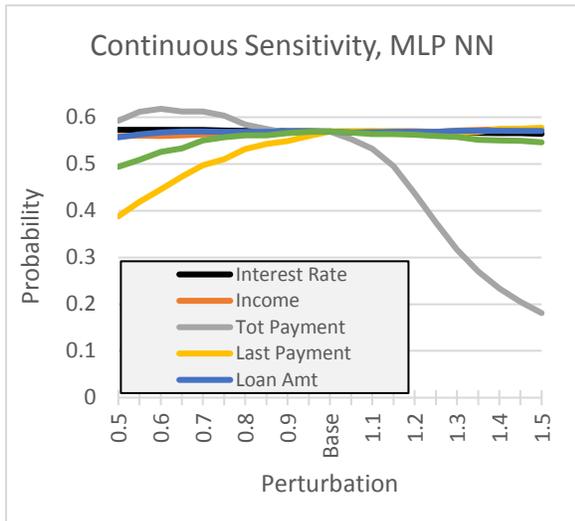

**Figure 6B. MLP NN prediction sensitivities for perturbed continuous variables.**

**Table 4. Perturbation table for continuous features** *int, ann_incme, total_pymnt*

|  |  | Predicted PROB | | | | |
|---|---|---|---|---|---|---|
|  |  | 0.8 | 0.9 | Base | 1.1 | 1.2 |
| int | rf | 0.2525 | 0.339 | 0.538 | 0.554 | 0.548 |
|  | GBC | 0.3015 | 0.208 | 0.532 | 0.4325 | 0.3525 |
|  | Logit | 0 | 0.0015 | 0.5865 | 0.948 | 0.988 |
|  | NN | 0.589 | 0.5875 | 0.5865 | 0.5855 | 0.5845 |
| ann_incme | rf | 0.5385 | 0.5415 | 0.538 | 0.5435 | 0.5435 |
|  | GBC | 0.5305 | 0.532 | 0.532 | 0.535 | 0.5335 |
|  | Logit | 0.5865 | 0.5865 | 0.5865 | 0.587 | 0.5885 |
|  | NN | 0.583 | 0.5875 | 0.5865 | 0.587 | 0.587 |
| total_pymnt | rf | 0.5735 | 0.553 | 0.538 | 0.5085 | 0.466 |
|  | GBC | 0.5305 | 0.5295 | 0.532 | 0.516 | 0.4855 |
|  | Logit | 0.619 | 0.6035 | 0.5865 | 0.5725 | 0.55 |
|  | NN | 0.607 | 0.5885 | 0.5865 | 0.5485 | 0.442 |

As before (and not surprisingly), the logit sensitivities conform with expectations. Model predictions are more sensitive to *Loan Grade A* and *Loan Grade D* (both in top 10 in terms of feature importance) than to *LoanTitle_CC* (which is not in the top-15 of actual feature importance).

In Figure 7B, we see that the RF model is very insensitive to changes in the categorical variables. Given SHAP-scores for loan-types A and D, this is surprising.

The GBC and MLP NN are even more insensitive—so the charts for these two classifiers have been eliminated—as no useful information is provided.

**Table 5. Perturbation table for categorical features** *Loan Type A*, *Loan Type D*, *Loan_Title_CC*

|  |  | Prediction Probability | | | | |
|---|---|---|---|---|---|---|
|  |  | Base (0) | 0.25 | 0.5 | 0.75 | 1 |
| Loan A | logit | 0.5785 | 0.6026 | 0.62614 | 0.6499 | 0.67436 |
|  | rfc | 0.545 | 0.55004 | 0.55532 | 0.55962 | 0.5644 |
|  | gbc | 0.522 | 0.52268 | 0.52304 | 0.52398 | 0.52444 |
|  | keras_ann | 0.5765 | 0.5765 | 0.5765 | 0.5765 | 0.5765 |
| Loan D | logit | 0.5785 | 0.54756 | 0.51598 | 0.48508 | 0.45422 |
|  | rfc | 0.545 | 0.5413 | 0.5369 | 0.53324 | 0.52916 |
|  | gbc | 0.522 | 0.52182 | 0.52168 | 0.52116 | 0.52138 |
|  | keras_ann | 0.5765 | 0.57642 | 0.5763 | 0.57636 | 0.57626 |
| Loan Title | logit | 0.5785 | 0.57912 | 0.57944 | 0.58014 | 0.58062 |
|  | rfc | 0.545 | 0.545 | 0.54512 | 0.54512 | 0.54526 |
|  | gbc | 0.522 | 0.52212 | 0.52202 | 0.52224 | 0.52216 |
|  | keras_ann | 0.5765 | 0.5765 | 0.5765 | 0.5765 | 0.5765 |

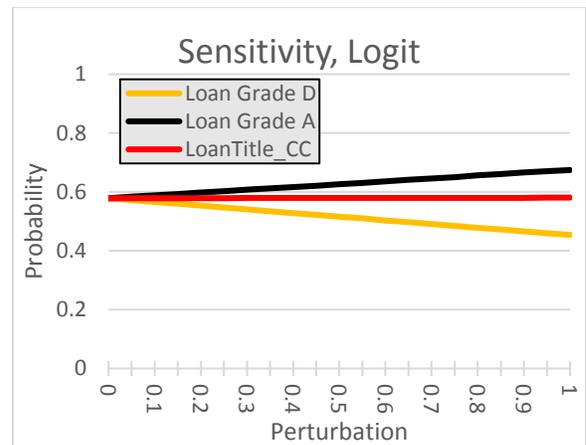

**Figure 7A. Logit prediction sensitivity for perturbed categorical features.**

---

[3] For the 'perturbation process' for loan-type. We increase the number of 1's by 5% (randomly selected) and change other associated loan-type (for the new 1's to be 0). We do this replacement 25 times and average the 25 outcomes. Now we increase the number of 1's by 5% (again) and repeat the process. We do this until we arrive at twice the original number of 1's in the test sample. Hence, in the charts, 1 indicates that we effectively doubled the number of 1's, while .5 indicates that we have increased the number of 1's by 50%.

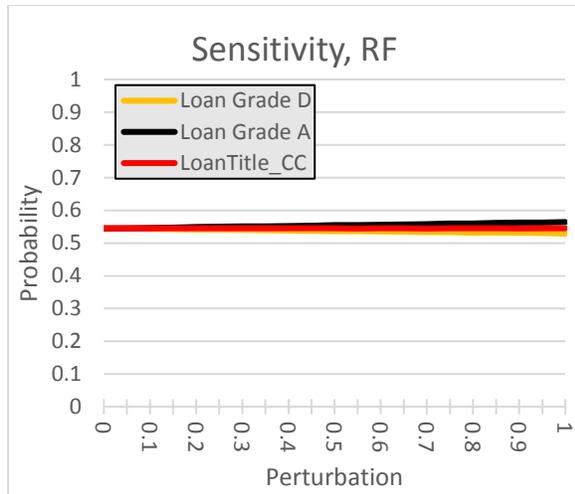

**Figure 7B. Random forest (RF) prediction sensitivity for perturbed categorical features.**

To summarize, we see that the sensitivities with respect to the continuous variables to be broadly consistent with expectations. (This is especially true for the logit model—which is to be expected.) For the other classifiers, there were some inconsistencies (either in relative sensitivity or in the projections themselves). Other than the logit model, the MLP NN model yielded predictions most in line with expectations gleaned from the explainability measures.

Likewise, the sensitivity of logit model predictions to perturbations in the categorical features follows expectations, e.g., loan-type D has a somewhat larger impact than does loan-type A. The other models yielded essentially unchanged estimates for changes in the categorical variables.

## 6. Summary

Explainable artificial intelligence (XAI) is a current research thrust devoted to demystifying "black box" models, especially involving neural networks. In this paper, we have addressed a subset of XAI, namely explaining and interpreting prediction models. In our example, we are interested in explaining a binary decision regarding credit card applications, whether to approve or deny an application. When talking about explainability, we have to ask "explainable to whom". In the latter case it is essential to present a coherent explanation to the applicant of why the credit card application was turned down. However, applicants are not the only stakeholders; corporate interests also must weigh the risk of defaults against the potential revenue stream of issuing new credit cards.

We have applied a portfolio of explanation techniques (LIME, SHAP, GAM) to determine which features have the biggest impact on this decision for a suite of different prediction models. These methods allow us not only a mechanism for comparing different prediction models but also provide significantly improved insight into the workings of models both at the local and the global levels. However, our work suggests that complex model explainability methods are still in the nascent stage for some real world deployment use cases such as credit denial explanations. Teasing a consensus from the portfolio of these techniques across multiple models is not always straightforward and can become an extended exercise in tradeoff analysis.

Our contribution has been to reveal a discontinuity between the static and dynamic explainability models which to our knowledge has not been identified in previous research. What we conclude from our experiment and suggest as future research are the following:

- Preliminary explainability prediction models provide a distinct improvement over the "black box".
- Extend the portfolio of prediction models (to include at a minimum SVM, Bayesian classifiers and additional N/N) and explainability techniques (to include at a minimum DeepLIFT and Integrated Gradients) to be analyzed and compared.
- Determining Feature Importance requires sophisticated statistical inference expertise and thus currently appears to be more useful to data scientists than to end users. Although Feature Importance charts have an intuitive appeal, more detailed analyses, Shapley diagrams for example, are not intuitive and need to be aggregated for better comprehensibility.
- This reveals a need for an Explainability DSS for decision makers that can integrate predictive modeling techniques with the explainability models associated with each. Requirements for such a DSS constitute a promising area of further research.
- More research is needed to understand and align prediction and base case feature importance incongruence.